\documentclass[11pt,a4paper]{article}
\usepackage[hyperref]{emnlp2020}
\usepackage{times}
\usepackage{latexsym}

\usepackage{microtype}
\usepackage{xcolor}

\aclfinalcopy % Uncomment this line for the final submission
\usepackage{tikz}
\usetikzlibrary{matrix,chains,positioning,decorations.pathreplacing,arrows}
\usepackage{float}
\usepackage{amssymb}
\usepackage{amsmath}
\usepackage{amsthm}
\usepackage{listings}
\usepackage{graphicx}
\graphicspath{ {./imgs/} }
\usepackage{caption}        % captioning in subfigures
\usepackage{sidecap}        % SCfigure
\usepackage{subcaption}
\usepackage[normalem]{ulem}

\title{CLEVR Parser: A Graph Parser Library for Geometric Learning on Language Grounded Image Scenes}

\author{Raeid Saqur\textsuperscript{\rm{1,2}} \\
\textsuperscript{\rm 1}University of Toronto Computer Science \\
\textsuperscript{\rm 2}Vector Institute for Artificial Intelligence \\
  \texttt{raeidsaqur@cs.toronto.edu} \\
  \And
  Ameet Deshpande \\
  Department of Computer Science\\
  Princeton University\\
  \texttt{asd@cs.princeton.edu} \\}
\date{}

\begin{document}
\maketitle
\begin{abstract}
The CLEVR dataset has been used extensively in language grounded visual reasoning in  Machine Learning (ML) and Natural Language Processing (NLP) domains. We present a \textbf{graph parser library} for CLEVR, that provides functionalities for object-centric attributes and relationships extraction, and construction of structural graph representations for dual modalities. Structural order-invariant representations enable geometric learning and can aid in downstream tasks like language grounding to vision, robotics, compositionality, interpretability, and computational grammar construction. We provide three extensible main components -- \textbf{parser, embedder, and visualizer} that can be tailored to suit specific learning setups. We also provide out-of-the-box functionality for seamless integration with popular deep graph neural network (GNN) libraries. Additionally, we discuss downstream usage and applications of the library, and how it accelerates research for the NLP research community\footnote{Code is available at - \url{https://github.com/raeidsaqur/clevr-parser}}.  
\end{abstract}

% ==============================================================================
\section{Introduction}\label{intro}
% ==============================================================================
 The CLEVR dataset \cite{johnson2017clevr} is a modern 3D incarnation of historically significant shapes-based datasets like SHRDLU \cite{winograd1970shrdlu}, used for demonstrating AI efficacy on language understanding \cite{ontanon2018shrdlu, winograd1980does, Hudson2018}. Although originally aimed at the visual question answering (VQA) problem \cite{Santoro2017, Hu2018}, its versatility has seen its use in diverse ML domains, including extensions to physics simulation engines for language augmented hierarchical reinforcement learning \cite{jiang2019language} and causal reasoning \cite{yi2019clevrer}.  

\begin{figure}[H]
    \centering
    \begin{subfigure}[c]{0.49\textwidth}
        \centering
        %\caption{Textual question parsed representation}
        \caption{Question on image (Figure \ref{fig:clevr-val-img}): `Is the color of the \textit{metal block} that is \textit{right} of the \textit{yellow rubber object} the same as the \textit{large metal cylinder}?'}
        \begin{tikzpicture}[thick, scale=0.6, every node/.style={scale=0.6}, 
            roundnode/.style={circle, draw=red!60, fill=red!5, very thick, minimum size=5mm},
            attrnode/.style={rectangle, draw=red!60, fill=red!5, very thick, minimum size=5mm, scale=0.8},
            objnode/.style={draw, shape=circle, fill=green, minimum size=7mm, scale=0.8}
            ]
            % \tikzstyle{every node}=[draw, shape=circle];
            %obj0
            \node[objnode] (obj0)  {$obj_1$};
            \node[attrnode] [left=of obj0] (a01)  {block};
            \node[attrnode] [right=of obj0] (a02) {metal};
            \draw [<-] (obj0) -- (a01); \draw [<-](obj0) -- (a02);
            %obj2
            \node [objnode] [above=of obj0] (obj1) {$obj_2$};
            \node [attrnode] [above left=of obj1] (a11) {object};
            \node [attrnode] [above=of obj1] (a12) {yellow};
            \node [attrnode] [above right=of obj1](a13) {rubber};
            \draw [<-] (obj1) -- (a11); \draw [<-] (obj1) -- (a12); \draw [<-] (obj1) -- (a13);
            %%obj3
            \node [objnode] [below=of obj0] (obj2) {$obj_3$};
            \node [attrnode] [below right=of obj2] (a21) {cylinder};
            \node [attrnode] [below=of obj2] (a22) {metal};
            \node [attrnode] [below left=of obj2](a23) {large};
            \draw [<-] (obj2) -- (a21); \draw [<-] (obj2) -- (a22); \draw [<-] (obj2) -- (a23);
            % obj-obj connections
            \draw [dashed] (obj0) -- (obj1) node[midway, left]{\texttt{spatial\_re}}; 
            \draw [dashed](obj0) -- (obj2) node[midway, left]{\texttt{matching\_re}};
        \end{tikzpicture}
        \label{fig:Gs-tikz}
    \end{subfigure}
    \vfill \vspace{6pt}
    \begin{subfigure}[c]{0.4\textwidth}
        \centering
        \caption{Image (Figure \ref{fig:clevr-val-img}) scene graph parsed representation}
        \includegraphics[width=0.9\linewidth]{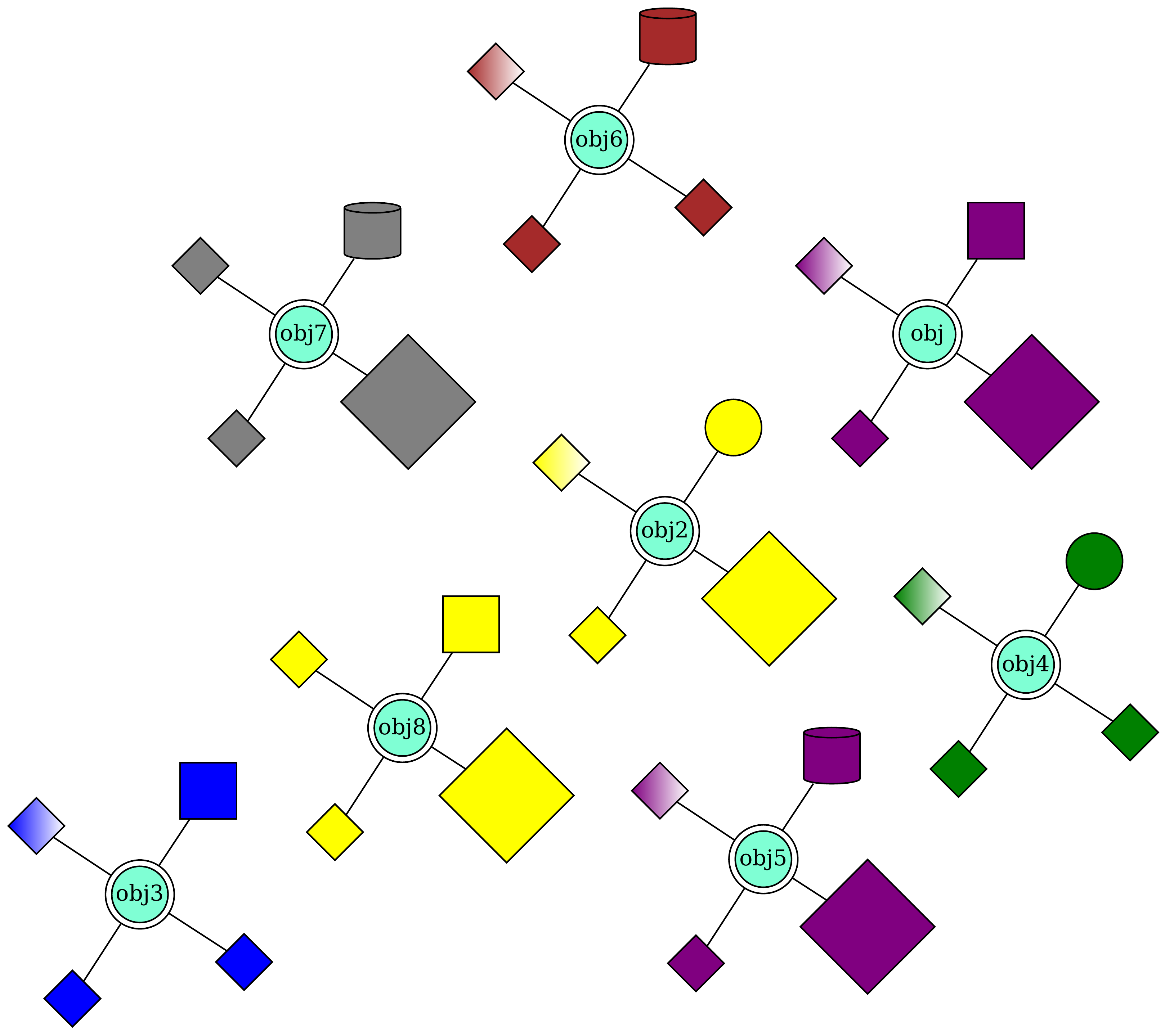}        
        \label{fig:Gt-graphviz}
    \end{subfigure} 
    \caption{A question about a CLEVR image visualized as multimodal parsed graphs}
    \label{fig:intro-visualization}
\end{figure}

Parallelly, research interest in geometric learning and GNN \cite{kipf2016semi, schlichtkrull2018modeling, hamilton2017representation} based techniques have seen a dramatic surge in recent deep learning zeitgeist. In this focused paper, we present a library that allows easy integration and application of geometric representation learning on CLEVR dataset tasks - enabling the NLP research community to apply GNN based techniques to their research (see \ref{sec:related-work}).

The library has three main (extensible) components: 
1. \textbf{Parser}: allows extraction of graph structured relationships among objects of the environment -- both for textual questions, and semantic image scene graphs, 2. \textbf{Embedder}: allows generation of latent embeddings using any models or desired backend of choice (like PyTorch\footnote{https://pytorch.org/}), 3. \textbf{Visualizer}: provides tools for visualizing structural graphs and latent embeddings.  
% \begin{enumerate}
%     \item \textbf{Parser}: allows extraction of graph structured relationships among objects of the environment -- both for textual questions, and semantic image scene graphs
%     \item \textbf{Embedder}: allows generation of latent embeddings using any models or desired backend of choice (like PyTorch\footnote{https://pytorch.org/}, Tensorflow\footnote{https://www.tensorflow.org/})
%     \item \textbf{Visualizer}: provides tools for visualizing structural graphs and latent embeddings
% \end{enumerate}

%Thus, with the release of this library, we hope to enable greater adoption of geometric learning in the NLP community, by lowering initial learning curve and/or rapid prototyping and integration of GNNs in NLP research domains like language grounded RL, visual reasoning, language compositionality \cite{bahdanau2019closure} etc. .

% ==============================================================================
\section{Background} \label{sec:background}
% ==============================================================================

\paragraph{CLEVR Environment} The dataset consists of images with rendered 3D objects of various shapes, colors, materials, and sizes, along with corresponding image scene graphs containing visual semantic information. Templated question generation on the images allows the creation of complex questions that test various aspects of scene understanding. The original dataset contains $\approx$1M questions generated from $\approx$100k questions with 90 question template families that can be broadly categorized into five question types: count, exist, numerical comparison, attribute comparison, and query. 

\begin{figure}[!htbp]
    \centering
    \includegraphics[width=0.9\linewidth]{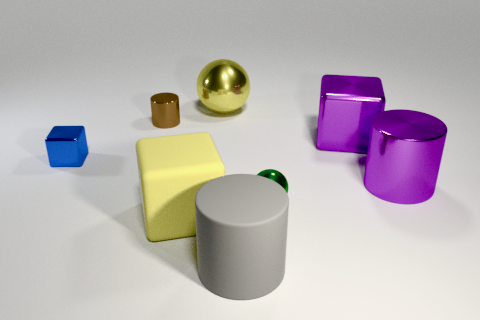}
    \caption{A CLEVR image}
    \label{fig:clevr-val-img}
\end{figure}

The dataset also comes with a defined domain-specific-language (DSL) function library $\mathcal{F}$, containing primitive functions that can be composed together to answer questions on CLEVR images \cite{johnson2017inferring}. We delegate further details of this dataset to \cite{johnson2017clevr} and the appendix \ref{sec:appendix}.

% ==============================================================================
\section{CLEVR-PARSER}\label{sec:clevr-parser}
% ==============================================================================
Here we describe each of the main library components in detail.
 
\subsection{Parser} \label{structural-parser}
\begin{figure}[!htbp]
    \centering
    \includegraphics[width=0.99\linewidth]{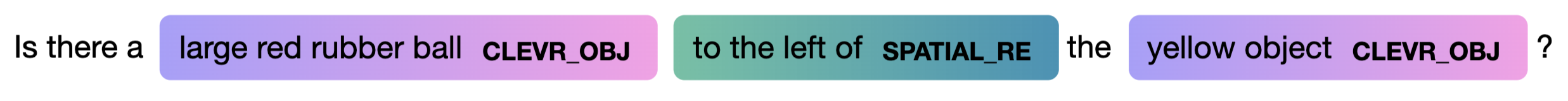}
    \caption{Entity visualization}
    \label{fig:entityviz}
\end{figure}

\paragraph{Text} The parser takes a language utterance, which can be a question, caption or command, that is valid in the CLEVR environment, and outputs a structural graph representation -- $G_s$, capturing object attributes, spatial relationships (\texttt{spatial\_re}), and attribute similarity based matching predicates (\texttt{matching\_re}) in the textual input. This is implemented by adding a CLEVR object entity recognizer (NER) in the NLP parse pipeline as depicted by Figure \ref{fig:entityviz}. Note that the NER is permutationally equivariant to the object attributes -- i.e. a `large red rubber ball' will be detected as an object by any of these spans: `red large rubber ball', `large ball', `ball' etc. 

\paragraph{Images} 
%For parsing images, it is presumed that the parsed semantic image scenes are available. 
The parser takes image scene graphs as input and outputs a structural graph -- $G_t$. The synthesized image scenes accompanying the original dataset can be used as input. Alternatively, parsed image scenes generated using any modern semantic image segmentation method (for e.g. `Mask-RCNN' \cite{he2017mask}) can also be used as input \cite{Yi2018}. A visualized example of a parsed image is shown in figure \ref{fig:Gt_vis}. For the ease of reproducibility, we also include a curated dataset `1obj' with parsed image scenes using Mask-RCNN semantic segmentation (Appendix\ref{sec:appendix}). 

While we provide a concrete implementation using the SpaCy\footnote{https://spacy.io/} NLP library, any other library like the Stanford Parser\footnote{https://nlp.stanford.edu/software/lex-parser.shtml}, or NLTK\footnote{https://www.nltk.org/} could be used in its place. The output of the parser from a question and image is depicted in Figure \ref{fig:intro-visualization}.
  
\subsection{Embedder}\label{embedder}
The embedder provides `word-embedding' \cite{mikolov2017advances} based representation of input text utterances and image scenes using a pre-trained language model (LM). The end-user can instantiate the embedder with a preferred LM, which could be a simple one-hot representation of the CLEVR environment vocabulary, or a large transformer based SotA LMs like BERT, GPT-2, XLNet \cite{peters2018deep, devlin2018bert, radford2019language, yang2019xlnet}. The embedder uses the parser (see section \ref{structural-parser}) generated graphs $\mathcal{G}_s, \mathcal{G}_t$ -- where graph $\mathcal{G}_s$ and $\mathcal{G}_t$ are defined as generic graph $\mathcal{G}$ = $(\mathcal{V}, \mathcal{E}, \mathcal{A} )$, where $\mathcal{V}$ is the set of nodes \{1,2,..\}, $\mathcal{E}$ is the set of edges, and $\mathcal{A}$ is the adjacency matrix -- and returns $\mathcal{X}$, $E$, the feature matrices of the nodes and edges respectively:

\begin{equation}
    \begin{aligned}
        \mathcal{X}_s, A_s, E_s \leftarrow \texttt{EMBED}(S)     \\
        \mathcal{X}_t, A_t, E_t \leftarrow \texttt{EMBED}(T)    ,
    \end{aligned}    
\end{equation}
%takes text tokens (S), image scene graphs (T), and outputs tuple ($\mathcal{X}$, A, E):
The output signature of the embedder is a tuple: ($\mathcal{X}$, A, E), which matches the fundamental data-structure of popular geometric learning libraries like PyTorch Geometric~\cite{fey2019fast}, thus allowing seamless integration. We show a concrete implementation of this use case using \textbf{PyTorch Geometric} \cite{fey2019fast} and Pytorch in \ref{viz-embd}.

\subsection{Visualizer}\label{visualizer}

We provide multiple visualization tools for analyzing images, text, and latent embeddings.

\subsubsection{Visualizing Structural Graphs}\label{viz-structural}
This visualizer sub-component enables visualization of the multimodal structural graph outputs -- $G_s, G_t$ -- by the parser (see \ref{structural-parser}) using \texttt{Graphviz} and \texttt{matplotlib}. 
%Both a CLEVR image and the corresponding question or caption can be viewed as a structural graph where attributes are connected to its object, and edges between objects encode their relations. 

\begin{figure} [h!]
    \centering    
    \begin{subfigure}[c]{0.49\textwidth}
        \includegraphics[width=0.9\linewidth]{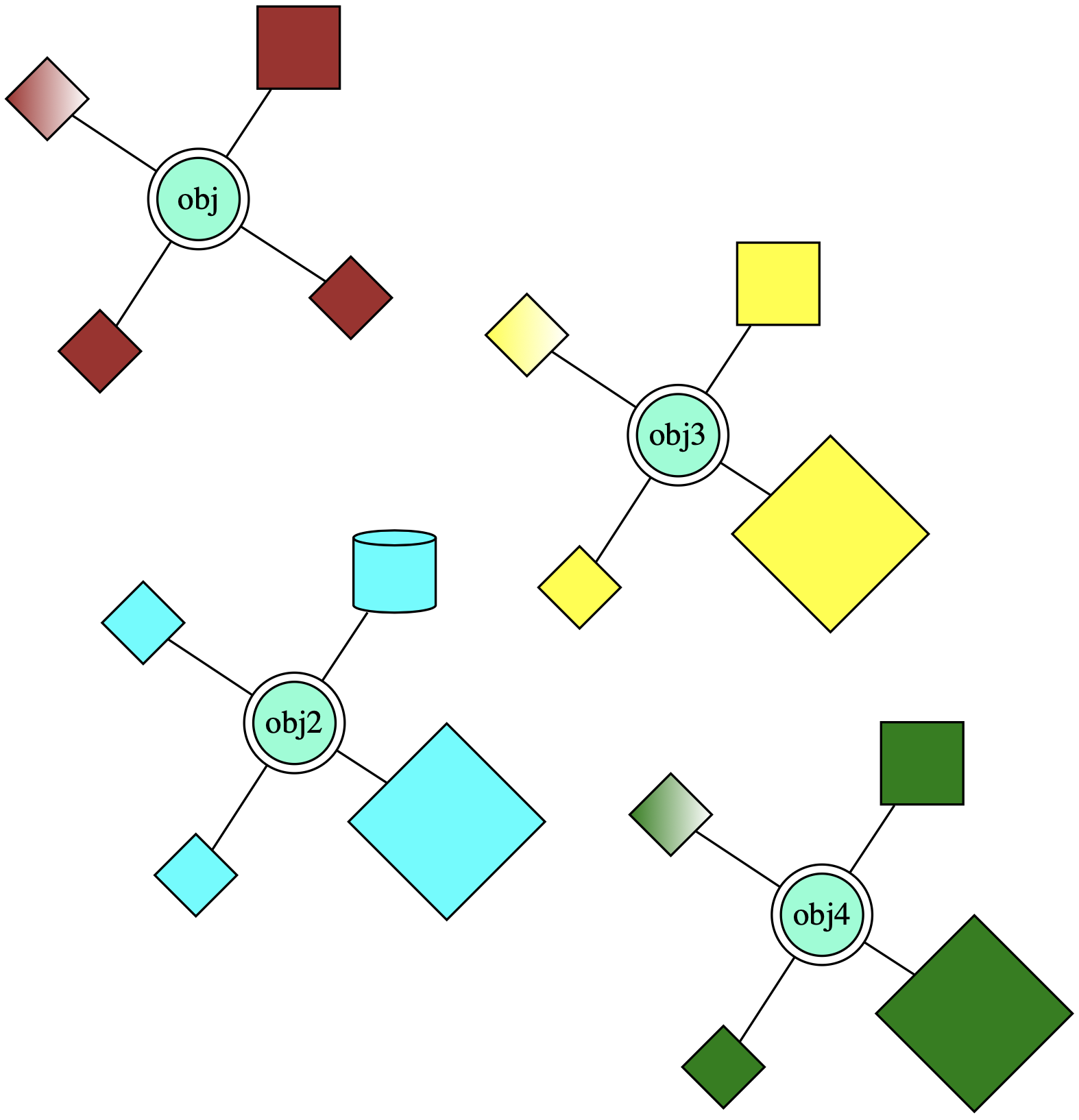}
        \caption{Visualizing image graph -- $G_t$}
        \label{fig:Gt_vis}
    \end{subfigure}
    \begin{subfigure}[c]{0.49\textwidth}
        \includegraphics[width=0.9\linewidth]{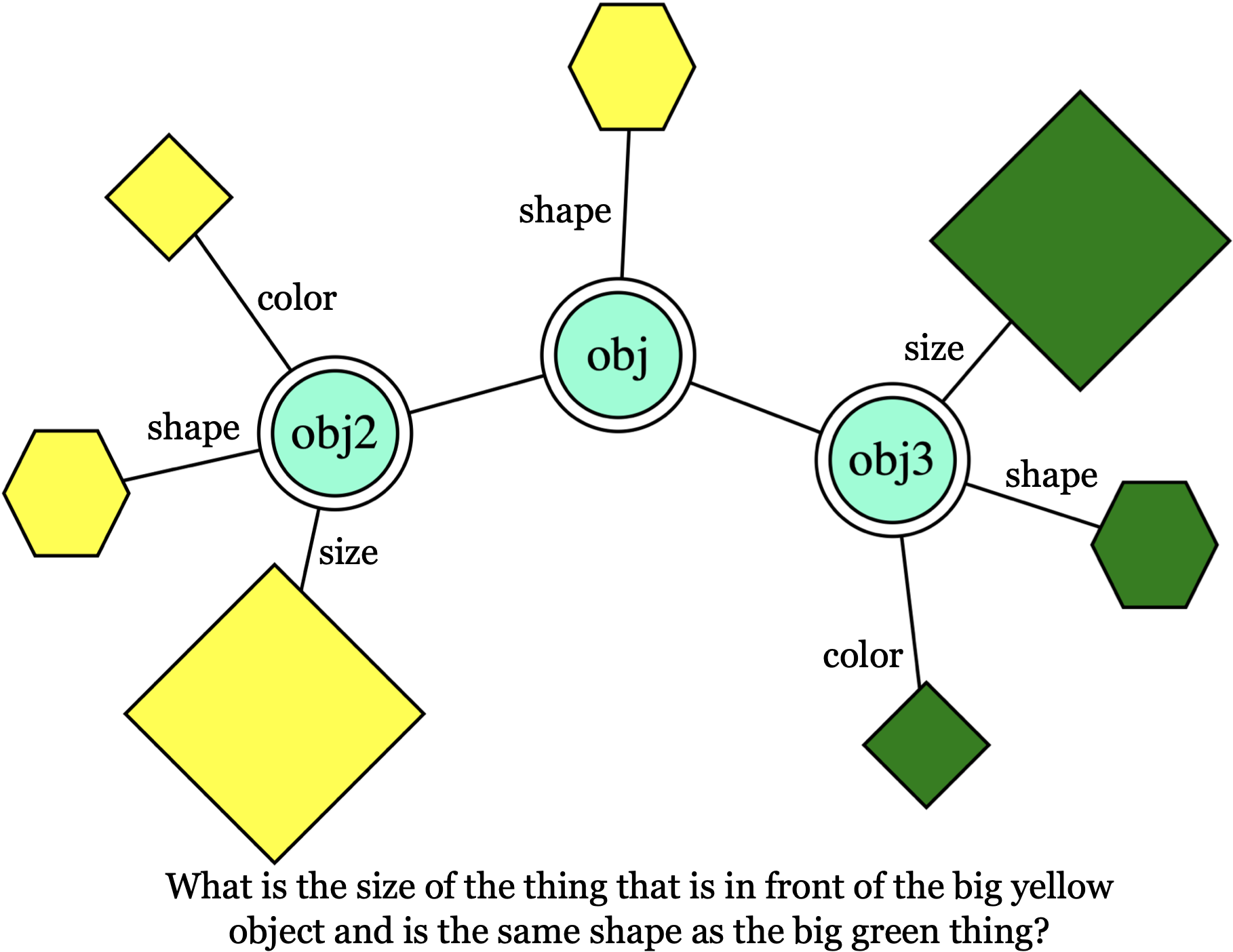}
        \caption{Visualizing text graph -- $G_s$}
        \label{fig:Gs_vis}
    \end{subfigure}
    \begin{subfigure}[c]{0.49\textwidth}
        \includegraphics[width=0.9\linewidth]{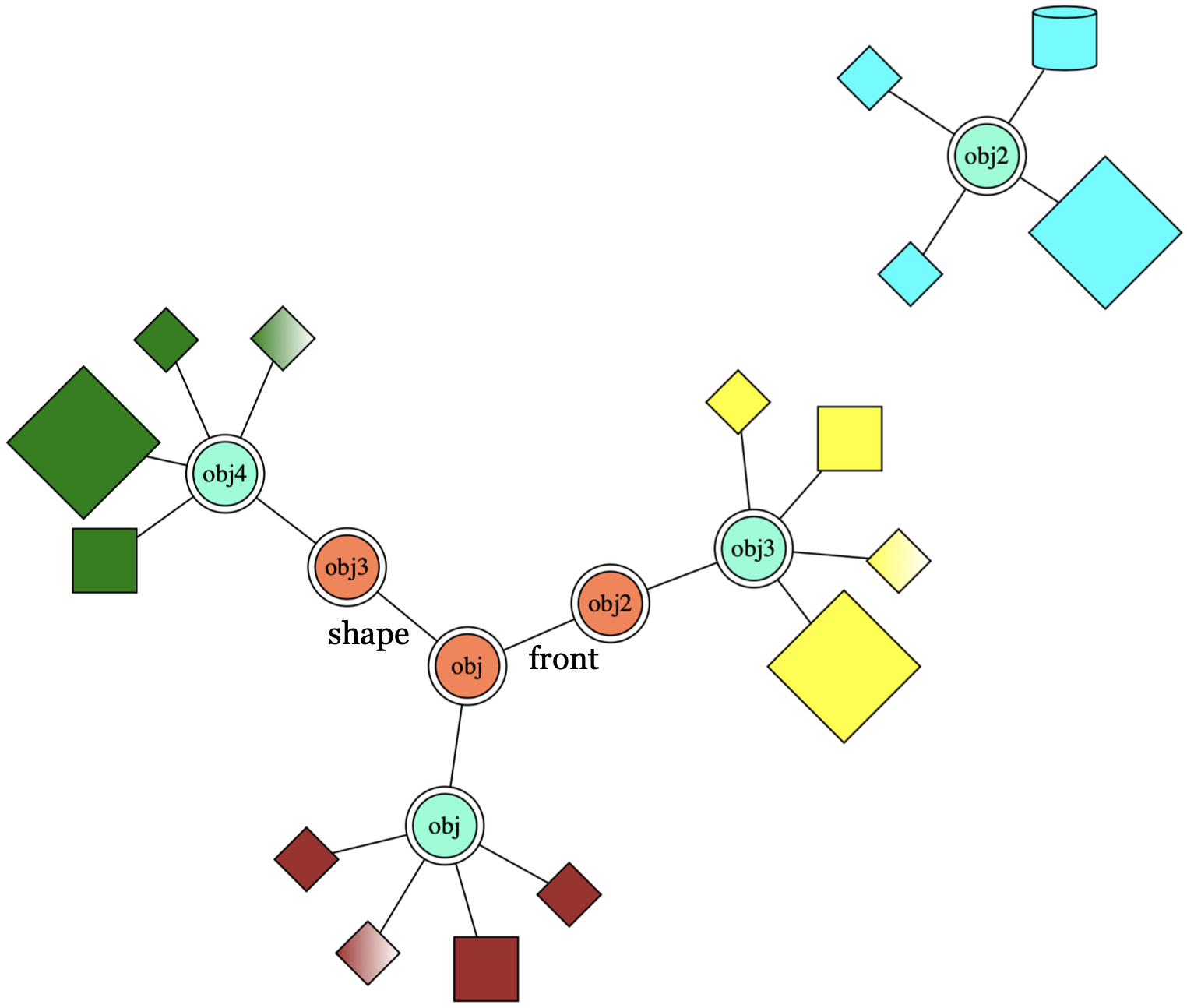}
        \caption{Visualizing joint (image and text) graph -- $G_u$ for the above two figures}
        \label{fig:Gu_vis}
    \end{subfigure}
    \caption{Visualizing $G_s, G_t, G_u$}    
\end{figure}

\paragraph{Visualizing Images} Image graphs ($G_t$) can have a large number of objects and attributes. For ease of viewing, attributes like size, shape (e.g. cylinder), color (e.g. yellow), and material (e.g. metallic) are displayed as nodes of the graph (Figure \ref{fig:Gt_vis}).
We explain elements of Figure~\ref{fig:Gt_vis} to describe the \textbf{legend} in greater detail. The double circles represent the objects, and the adjacent nodes are their attributes. 
%CLEVR objects have $4$ attributes, shape, size, color, and texture.
The \textit{shape} is depicted using the actual shape (e.g. the cyan cylinder -- $obj2$), and the other attributes are depicted as diamonds. The \textit{size} of one of the diamonds depicts if the object is small or large, e.g. the large cyan diamond attached to $obj2$ means that it is large. The \textit{color} of all the attribute nodes depicts the color of the object (e.g. the cyan color of $obj2$). The presence of a gradient in the remaining diamond depicts the \textit{material} of the object. For example, the gradient in the diamond attached to $obj4$ means that it is \textit{metallic}, and the solid fill for $obj2$ means that it is \textit{rubber}. While this legend is a little lengthy, we found that it makes visualization easier, but the user can choose to revert to the simpler setting of using text to depict the attributes.
% \paragraph{Visualizing Images} Image graphs ($G_t$) can have a large number of objects and attributes. For ease of viewing, attributes like shape (e.g. cylinder), color (e.g. yellow), and texture (e.g. metallic) are displayed as node attributes on the graph (Figure \ref{fig:Gt_vis}). We explain elements of figure~\ref{fig:Gt_vis} to describe the \textbf{legend} in greater detail. The double circles in figure~\ref{fig:Gt_vis} represent the objects, and the nodes attached to them are their attributes. CLEVR objects have $4$ attributes, shape, size, color, and texture. The shape attribute is depicted using the actual shape, e.g. the cyan cylinder ($obj2$), and the other attributes are depicted as diamonds. The size of one of the diamonds depicts if the object is large, medium, or small, e.g. the large green diamond attached to $obj4$ means that $obj4$ is large. The color of all the attribute nodes depict the color of the object (e.g. the yellow color of $obj3$). The presence of gradient of the remaining diamond depicts the texture of the object. For example, the gradient in $obj$ means that it is \textit{metallic}, and the solid fill of $obj2$ means that it is \textit{rubber}. While this legend is a little lengthy, we found that it makes visualization easier, but the user can choose to revert to the simpler setting of using text to depict the attributes.
% %and our tool effectively depicts them with minimal clutter.

\paragraph{Visualizing Text} Text corresponding to an image is a partially observable subset of objects, their relationships, and attributes. The dependency graph of the text is visualized just like the images, with only the observable information being depicted (Figure \ref{fig:Gs_vis}).

\paragraph{Composing image and text} We also provide an option to view an image and the text in the same graph. By connecting corresponding object nodes from the image and text, we create a bipartite graph that allows us to visualize all the information that an image-text pair contains (Figure \ref{fig:Gu_vis}). Additional examples from the visualizer are presented in appendix~\ref{appendix:additionalsamples}.

%\RS{saving space}
%\begin{enumerate}
%    \item \textbf{Visualizing images} Images can typically have a large number of objects and attributes, and our tool effectively depicts all the information with less clutter. For ease of viewing, attributes like shape, color, and texture are all displayed on the graph without any text (Figure \ref{fig:Gt_vis}).
%    \item \textbf{Visualizing Text} Text corresponding to an image is partially observable, and encodes a subset of relationships between objects and their corresponding attributes. The dependency graph of the text is visualized just like the images, with only the observable information being depicted (Figure \ref{fig:Gs_vis}).
%    \item \textbf{Composing image and text} We also provide an option to view an image and the text in the same graph. By connecting corresponding object nodes from the image and text, we create a bipartite graph that allows us to visualize all the information that an image-text pair contains (Figure \ref{fig:Gu_vis}).
%\end{enumerate}

\subsubsection{Visualizer - Embeddings}\label{viz-embd}

We also provide a visualizer to analyze the embeddings produced by using methods in section \ref{embedder}. We use t-SNE \cite{maaten2008visualizing}, which is a method used to visualize high-dimensional data on 2 or 3 dimensions. We also offer clustering support to allow grouping of similar embeddings together. Both image ~\cite{frome2013devise} and word embeddings~\cite{mikolov2013distributed} from learned models have the nice property of capturing semantic information, and our visualizers capture this semantic similarity information in the form of clusters.

\begin{figure}[H]
    \centering
    \includegraphics[width=0.9\linewidth]{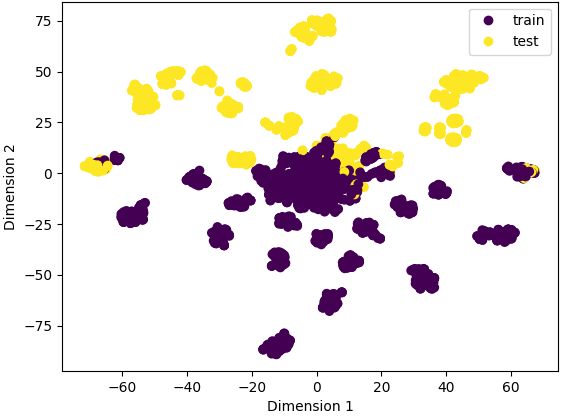}
    \caption{Questions from two different distributions which form separate clusters}
    \label{fig:2clusters}
\end{figure}

Figure \ref{fig:2clusters} plots the embeddings for questions drawn from two different distributions \textit{train} and \textit{test}, which represent semantically different sequences, and they separate out into distinct clusters. 

\begin{figure}[!htbp]
    \centering
    \includegraphics[width=0.9\linewidth]{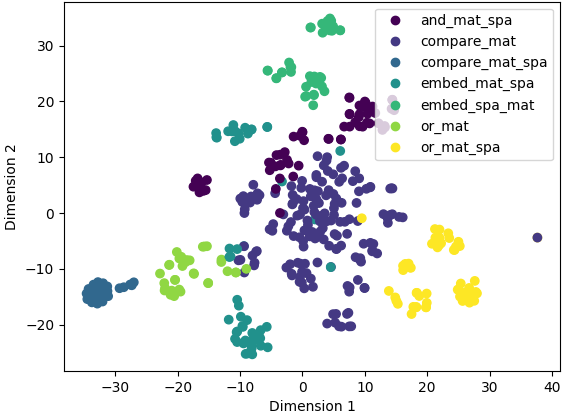}
    \caption{Questions from 7 different templates forming tight clusters}
    \label{fig:7clusters}
\end{figure}

Similarly, Figure \ref{fig:7clusters} analyzes embeddings drawn from 7 different templates. Questions that correspond to the same templates form tight clusters while being far away from other questions.

\section{Related Work and Applications} \label{sec:related-work}
%\ameet{Change section name}
Some lines of work attempt to generate scene graphs for images. The Visual Genome library \cite{krishna2017visual}, in a real-world image setting, is a collection of annotated images (from Flickr, COCO) and corresponding knowledge graph associations. The work of \cite{schuster2015generating} and the corresponding library which is a part of the Stanford NLP library\footnote{\url{https://nlp.stanford.edu/software/scenegraph-parser.shtml}}, allows scene graph generation from text (image caption) as input. 

Our work is orthogonal to these in that our target dataset is synthetic, which allows full control over the generation of images, questions, and ground truth semantic program chains. Thus, coalesced with our library's functionalities, it allows end-to-end (e2e) control over experimenting on every modular aspect of research hypotheses (see \ref{sec:applications}). Further, our work premises on providing multimodal representations -- including ground-truth paired graph (joint graph $G_u \leftarrow (G_s, G_t)$) -- which has interesting downstream research applications.

% \paragraph{Usages and Applications}\label{sec:applications}
\subsection{Usages and Applications}\label{sec:applications}
Applications of language grounding in ML/NLP research are quite broad. To avoid sounding overly grandiose, we exemplify possible applications citing work that pertains to the CLEVR dataset. 

Recent work by \cite{Bahdanau2019a} has shown lack of distributional robustness and compositional generalization \cite{fodor1988connectionism} in NLP. Permutation equivariance within local linguistic component groups has been shown to help with language compositionality \cite{gordon2020permutation}. Graph-based representations are intrinsically order invariant -- thus, may help with language compositionality research. 
Language augmented reward mechanisms are a dense topic in concurrent (human-in-the-loop) reinforcement learning \cite{knox2012reinforcement, griffith2013policy}, robotics \cite{knox2013training, kuhlmann2004guiding}, long-horizon, hierarchical POMDP problems in general \cite{kaplan2017beating} -- like command completion in physics simulators \cite{jiang2019language}. Other applications could be in program synthesis and interpretability \cite{mascharka2018transparency}, causal reasoning \cite{yao-2010-stage}, and general visually grounded language understanding \cite{yu2017training}.

In general, we expect and hope that any existing line or domain of work in NLP using the CLEVR dataset (a significant number), will benefit from having graph-based representational learning aided by our proposed library.

\bibliographystyle{acl_natbib}
\bibliography{anthology,emnlp2020,bibs/clevr}

% Submit appendix.pdf as a separate pdf file
\appendix
\section{Appendices} \label{sec:appendix}

\subsection{CLEVR Dataset}
\begin{figure}[!htbp]
\begin{center}
\includegraphics[width=0.49\textwidth]{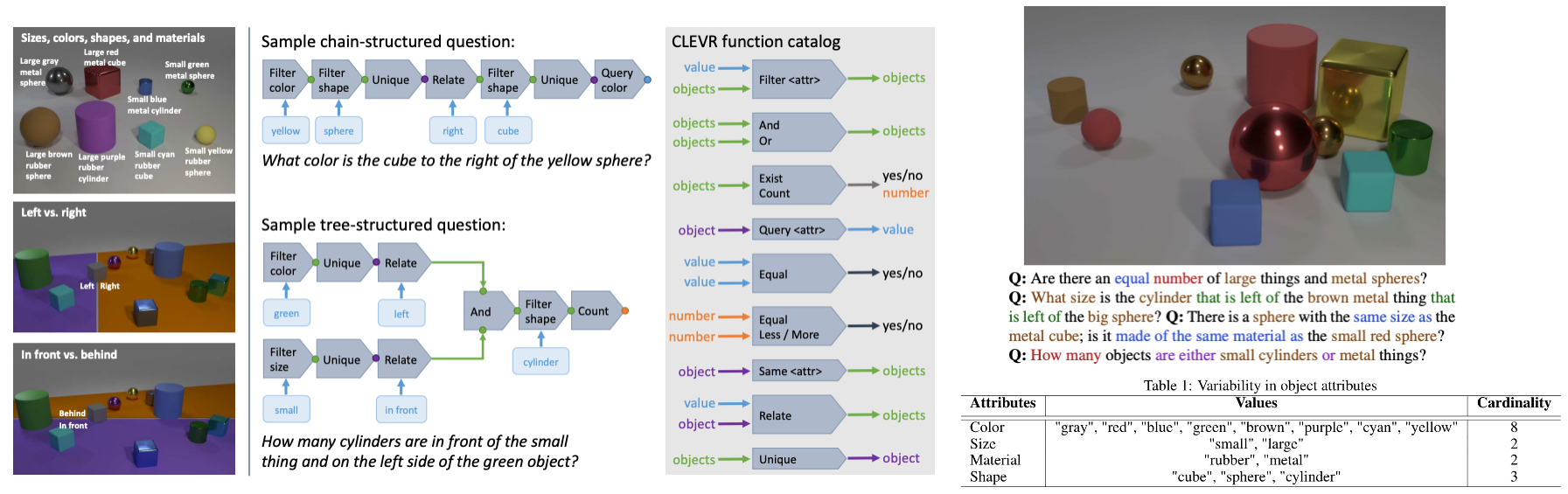}
\end{center}
\caption{Overview of the CLEVR dataset}\label{fig:clevr-dataset-overview}
\end{figure}

%\section{CLEVR Domain-Specific Language (DSL) and Functions Library} \label{appendix:dsl-library}
Figure \ref{fig:clevr-dataset-overview} shows a topological overview the dataset with sample image, corresponding questions, program chains using the function catalogue.

For detailed information about the function library accompanying the dataset release, please refer to CLEVR VQA \cite{johnson2017inferring}. The functions and signatures were kept unaltered from the original specifications \footnote{https://github.com/facebookresearch/clevr-dataset-gen/blob/master/question\_generation/metadata.json}.

%\begin{table}[H]
%\centering
%\caption{Variability in object attributes} \label{tab:clevr-obj-attrs}
%\resizebox{\textwidth}{!}{
%    \begingroup
%    \setlength{\tabcolsep}{6pt} % Default value: 6pt
%    \renewcommand{\arraystretch}{1.15} % Default value: 1
%        \begin{tabular}{| l | c | c | c |} % <-- alignment left, center, center.
%        \hline
%          \textbf{Attributes} & \textbf{Symbols} & \textbf{Values} & \textbf{Cardinality}\\ [0.5ex]
%          \hline \hline
%          Color & <C> & "gray", "red", "blue", "green", "brown", "purple", "cyan", "yellow" & 8 \\
%          Size  & <Z> & "small", "large" & 2 \\
%          Material & <M> & "rubber", "metal" & 2 \\
%          Shape & <S> & "cube", "sphere", "cylinder" & 3 \\
%          \hline
%        \end{tabular}
%    \endgroup
%}
%\end{table}

%\label{app:data}

\subsection{Data}
The regular CLEVR dataset can be downloaded from the project page of \cite{johnson2017clevr}. We also include a curated dataset`\textbf{1obj}' which contains synthetic data created for a single CLEVR object. In addition to synthetically created questions, images, image scenes, this dataset also contains the images parsed through a semantic image segmentation layer (Mask-RCNN) and a compositional dataset `1obj-CoGenT' which uses distributionally shifted object attributes for training and test data (for the compositional CoGenT test).

\subsection{Structural Graph Visualizations from Parser}

This section contains additional visualizations for the parser.

\begin{figure}[H]
    \centering
    \includegraphics[width=0.9\linewidth]{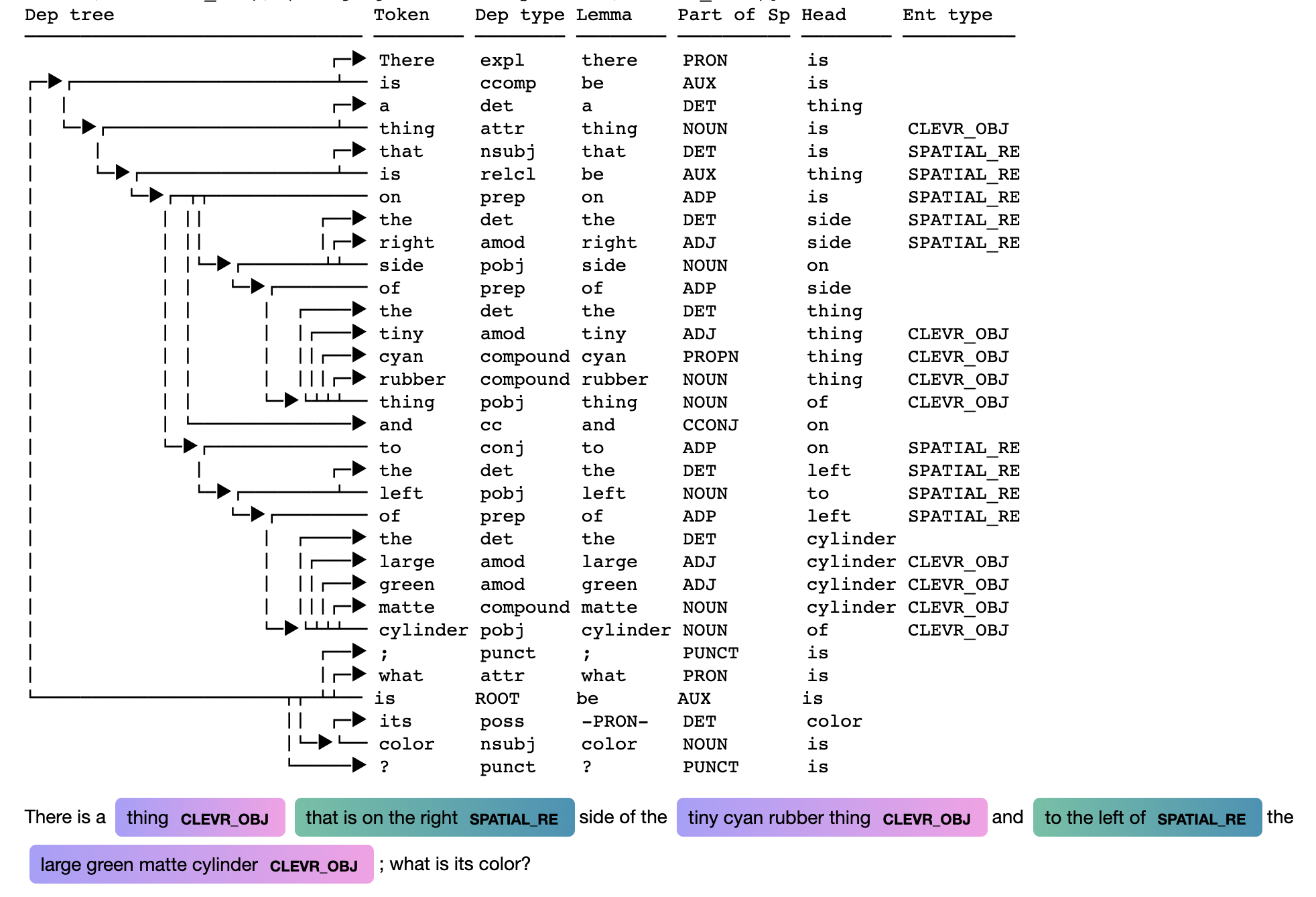}
    \caption{Entity visualization}
    \label{fig:entity-viz}
\end{figure}

% Submissions may include non-readable supplementary material used in the work and described in the paper.
% Any accompanying software and/or data should include licenses and documentation of research review as appropriate.
% Supplementary material may report preprocessing decisions, model parameters, and other details necessary for the replication of the experiments reported in the paper.
% Seemingly small preprocessing decisions can sometimes make a large difference in performance, so it is crucial to record such decisions to precisely characterize state-of-the-art methods. 

% Nonetheless, supplementary material should be supplementary (rather than central) to the paper.
% \textbf{Submissions that misuse the supplementary material may be rejected without review.}
% Supplementary material may include explanations or details of proofs or derivations that do not fit into the paper, lists of
% features or feature templates, sample inputs and outputs for a system, pseudo-code or source code, and data.
% (Source code and data should be separate uploads, rather than part of the paper).

% The paper should not rely on the supplementary material: while the paper may refer to and cite the supplementary material and the supplementary material will be available to the reviewers, they will not be asked to review the supplementary material.

\subsection{Additional Samples}\label{appendix:additionalsamples}

This section contains additional examples from the visualizer.
\begin{figure}[H]
    \centering
    \includegraphics[width=0.9\linewidth]{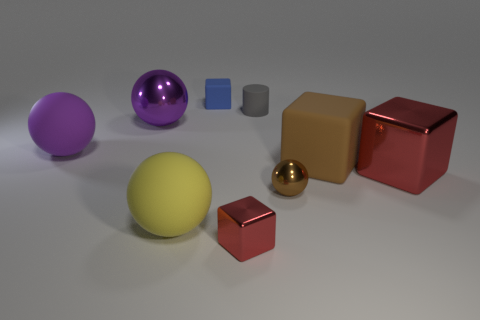}
    \caption{An Image from the CLEVR dataset}
    \label{fig:a4clevrimage}
\end{figure}

\begin{figure}[H]
    \centering
    \includegraphics[width=0.9\linewidth]{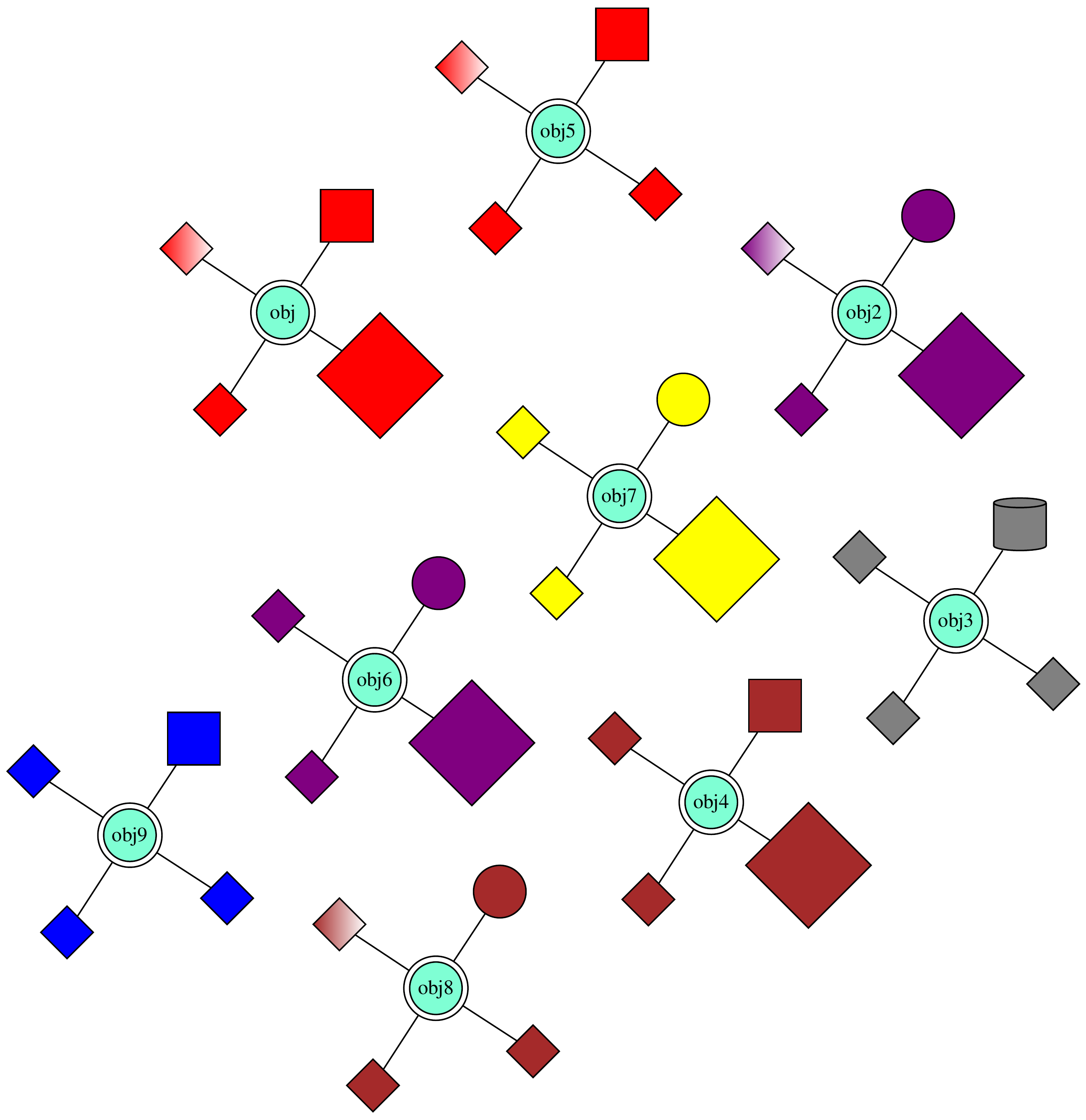}
    \caption{Corresponding structural graph representation of the image}
    \label{fig:a4gt}
\end{figure}

%% Fig 11 %%
\begin{figure}[H]
    \centering
    \includegraphics[width=0.9\linewidth]{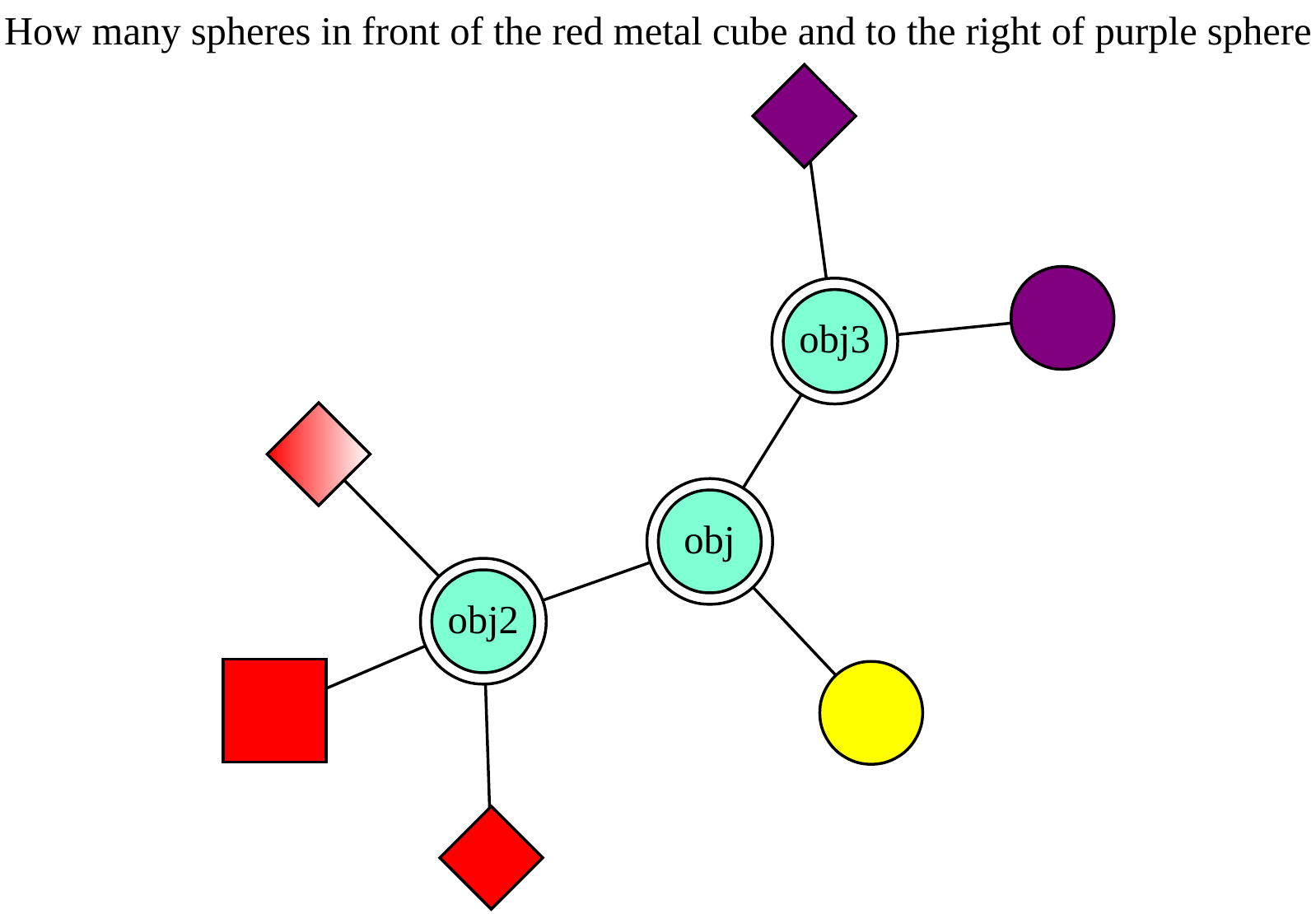}
    \caption{An example question for the corresponding image}
    \label{fig:a4gs}
\end{figure}

\begin{figure}[H]
    \centering
    \includegraphics[width=0.9\linewidth]{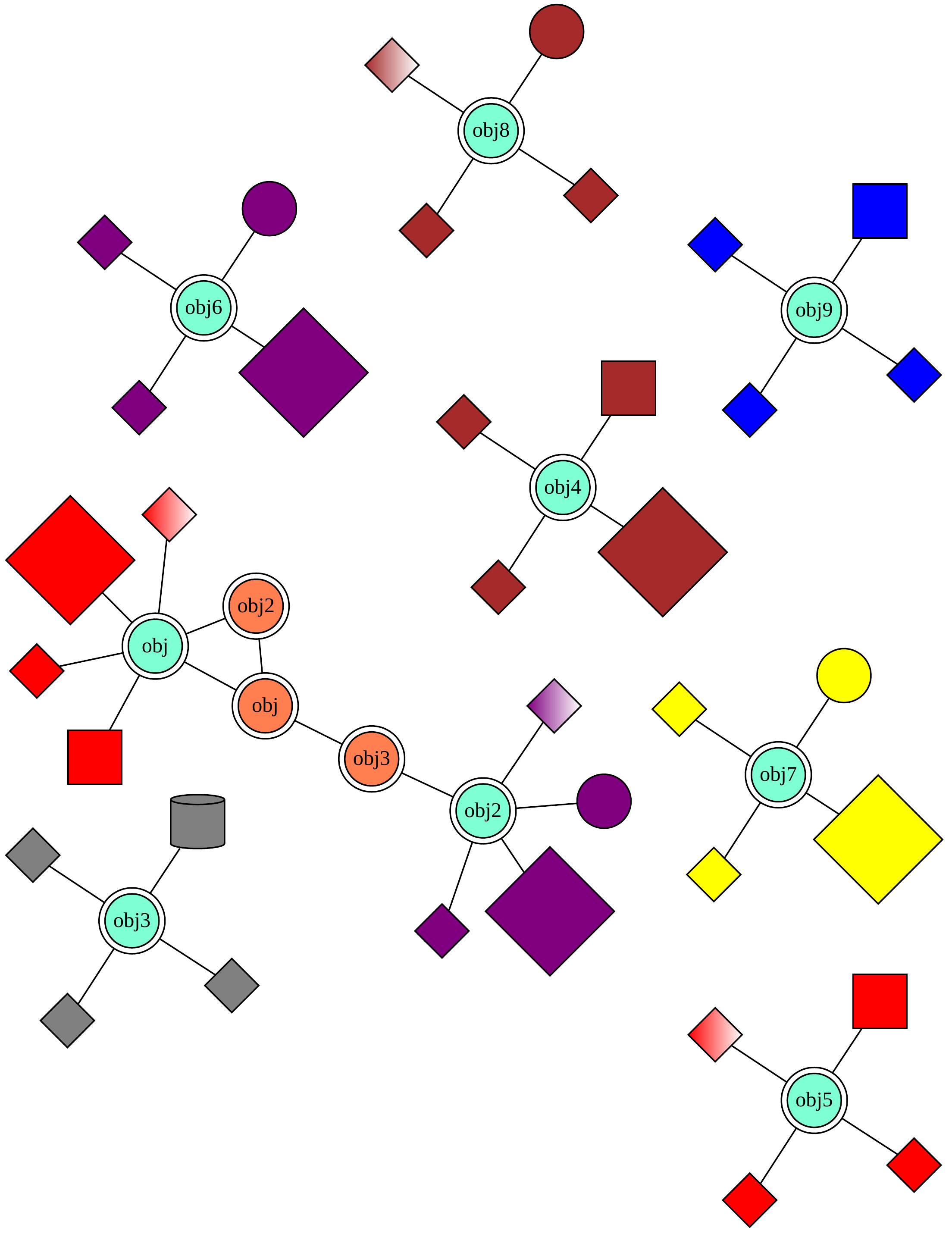}
    \caption{Visualizing both the image and the question in a bipartite graph. Orange nodes represent the text nodes, and the aquamarine nodes represent the image nodes}
    \label{fig:a4gu}
\end{figure}

\end{document}